\documentclass[conference]{IEEEtran}

\usepackage{amsmath, amsthm, amssymb}
\usepackage{graphicx}
\usepackage{verbatim}
\usepackage{caption}
\usepackage{subcaption}
\usepackage{fancyvrb}
\usepackage{enumerate}
\usepackage{relsize}
\usepackage{amsfonts}

\usepackage{algorithm}
\usepackage[noend]{algpseudocode}

\usepackage{tikz}
\usetikzlibrary{fit,positioning}

\usepackage{hyperref}

\makeatletter
\newcommand*\if@single[3]{%
  \setbox0\hbox{${\mathaccent"0362{#1}}^H$}%
  \setbox2\hbox{${\mathaccent"0362{\kern0pt#1}}^H$}%
  \ifdim\ht0=\ht2 #3\else #2\fi
  }
  
\newcommand*\rel@kern[1]{\kern#1\dimexpr\macc@kerna}

\newcommand*\widebar[1]{\@ifnextchar^{{\wide@bar{#1}{0}}}{\wide@bar{#1}{1}}}

\newcommand*\wide@bar[2]{\if@single{#1}{\wide@bar@{#1}{#2}{1}}{\wide@bar@{#1}{#2}{2}}}
\newcommand*\wide@bar@[3]{%
  \begingroup
  \def\mathaccent##1##2{%
    \if#32 \let\macc@nucleus\first@char \fi
    \setbox\z@\hbox{$\macc@style{\macc@nucleus}_{}$}%
    \setbox\tw@\hbox{$\macc@style{\macc@nucleus}{}_{}$}%
    \dimen@\wd\tw@
    \advance\dimen@-\wd\z@
    \divide\dimen@ 3
    \@tempdima\wd\tw@
    \advance\@tempdima-\scriptspace
    \divide\@tempdima 10
    \advance\dimen@-\@tempdima
    \ifdim\dimen@>\z@ \dimen@0pt\fi
    \rel@kern{0.6}\kern-\dimen@
    \if#31
      \overline{\rel@kern{-0.6}\kern\dimen@\macc@nucleus\rel@kern{0.4}\kern\dimen@}%
      \advance\dimen@0.4\dimexpr\macc@kerna
      \let\final@kern#2%
      \ifdim\dimen@<\z@ \let\final@kern1\fi
      \if\final@kern1 \kern-\dimen@\fi
    \else
      \overline{\rel@kern{-0.6}\kern\dimen@#1}%
    \fi
  }%
  \macc@depth\@ne
  \let\math@bgroup\@empty \let\math@egroup\macc@set@skewchar
  \mathsurround\z@ \frozen@everymath{\mathgroup\macc@group\relax}%
  \macc@set@skewchar\relax
  \let\mathaccentV\macc@nested@a
  \if#31
    \macc@nested@a\relax111{#1}%
  \else
    \def\gobble@till@marker##1\endmarker{}%
    \futurelet\first@char\gobble@till@marker#1\endmarker
    \ifcat\noexpand\first@char A\else
      \def\first@char{}%
    \fi
    \macc@nested@a\relax111{\first@char}%
  \fi
  \endgroup
}
\makeatother

\makeatletter
\newcommand{\vast}{\bBigg@{3}}
\newcommand{\Vast}{\bBigg@{4}}
\makeatother

\newcommand{\gcal}{\mathcal{G}}

\newcommand{\PP}[2][]{\mathbb{P}_{#1}\left[#2\right]}

\graphicspath{{./paper/figures/}}

\usepackage{lscape}

\usepackage{multicol}

\theoremstyle{plain}
\newtheorem{prop}{Proposition}

\newtheorem{lemm}[prop]{Lemma}

\theoremstyle{definition}

\theoremstyle{remark}

\begin{document}
\title{Efficient Representations for High-Cardinality Categorical Variables in Machine Learning}
\author{\IEEEauthorblockN{Zixuan Liang}
\IEEEauthorblockA{\textit{zliang1@my.harrisburgu.edu}}
}

\maketitle

\begin{abstract}
  High-cardinality categorical variables pose significant challenges in machine learning, particularly in terms of computational efficiency and model interpretability. Traditional one-hot encoding often results in high-dimensional sparse feature spaces, increasing the risk of overfitting and reducing scalability. This paper introduces novel encoding techniques, including means encoding, low-rank encoding, and multinomial logistic regression encoding, to address these challenges. These methods leverage sufficient representations to generate compact and informative embeddings of categorical data. We conduct rigorous theoretical analyses and empirical validations on diverse datasets, demonstrating significant improvements in model performance and computational efficiency compared to baseline methods. The proposed techniques are particularly effective in domains requiring scalable solutions for large datasets, paving the way for more robust and efficient applications in machine learning.

\end{abstract}

\section{Introduction}

Many regression problems often arise when analyzing data collected across several groups, each of which can contribute uniquely to the statistical relevance of the model. For instance, in a healthcare context, if I want to model health outcomes using patient data from various hospitals, I need to account for hospital-specific effects that other covariates might not capture. This type of group effect is observed in numerous contexts, such as examining students from different schools, voters living in separate zip codes, employees at various firms, and many more scenarios.

A widely employed method for addressing such issues is fixed effect modeling. To illustrate, let’s assume that I have $n$ samples $\{X_i, G_i, Y_i\}$ for $i = 1, \ldots, n$, where $X_i \in \mathbb{R}^p$ is a set of covariates pertaining to individual subjects, $G_i \in \mathcal{G}$ is a categorical variable denoting group membership, and $Y_i \in \mathbb{R}$ is the outcome of interest. My aim is to estimate:
\begin{equation}
\label{eq:mu}
\mu(x, g) = \mathbb{E}[Y_i \mid X_i = x, \, G_i = g].
\end{equation}
The conventional fixed effects approach models this expectation as:
\begin{equation}
\label{eq:FE}
\mu(x, g) = \alpha_g + x\beta,
\end{equation}
where the parameters $\beta$ and $\alpha_g$ are estimated through ordinary least squares regression. Extensions of this basic approach can involve applying non-linear transformations to $x$, considering interactions between the group membership variable and other covariates, or adding regularization to improve stability and predictive power \cite{angrist2008mostly,diggle2002analysis,wooldridge2010econometric}.

However, in practice, I often observe that fixed effects modeling faces limitations when the underlying signal is complex and non-linear or when the number of groups, $|\mathcal{G}|$, is large. The rigidity of the model \eqref{eq:FE} can hinder its ability to capture complex signals, and the high number of $\alpha_g$ parameters can lead to challenges with statistical inference \cite{neyman1948consistent}. In essence, the model might become too cumbersome to yield stable results while also lacking the flexibility to capture nuanced patterns in the data.

To address these challenges, the focus of this paper is to propose a more efficient way to represent group membership. Specifically, I seek to develop a mapping $\psi$ that transforms group membership $G_i$ into a $k$-dimensional space without compromising predictive capacity:
\begin{equation}
\label{eq:repr}
\psi: \mathcal{G} \rightarrow \mathbb{R}^k, \quad \mu(x, g) = f(x, \psi(g)),
\end{equation}
where $k$ is relatively small (i.e., $k \ll |\mathcal{G}|$) and the function $f(\cdot, \cdot)$ remains easy to learn. With such a mapping, the problem in \eqref{eq:mu} reduces to a standard regression problem with $(p+k)$-dimensional real-valued features $(X_i, \psi(G_i))$, allowing me to apply off-the-shelf statistical learning tools.

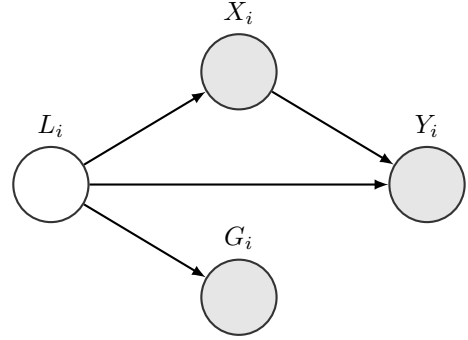
\begin{figure}[t]
\centering
\begin{tikzpicture}
\tikzstyle{main}=[circle, minimum size = 10mm, thick, draw=black!80, node distance = 14mm]
\tikzstyle{connect}=[-latex, thick]
\tikzstyle{box}=[rectangle, draw=black!100]
\node[main] (L) [xshift=-2.5cm, label=$L_i$] {};
\node[main, fill=black!10] (X) [yshift=1.5cm, label=$X_i$] {};
\node[main, fill=black!10] (G) [yshift=-1.5cm, label=$G_i$] {};
\node[main, fill=black!10] (Y) [xshift=2.5cm, label=$Y_i$] {};
\path
    (L) edge [connect] (X)
    (L) edge [connect] (G)
    (X) edge [connect] (Y)
    (L) edge [connect] (Y);
\end{tikzpicture}
\caption{This causal graph shows the core assumption that $Y_i$ and $X_i$ are conditionally independent of group membership $G_i$ when conditioned on latent state $L_i$. Gray nodes denote observed variables.}
\label{fig:graph_simple}
\end{figure}

To create an effective representation of group membership $G_i$, I must make assumptions about the link between $G_i$ and the outcome $Y_i$. The central premise, referred to as the \emph{sufficient latent state assumption}, is illustrated in Figure \ref{fig:graph_simple}. This assumption posits that while $G_i$ does not directly impact $Y_i$, it may relate to latent variables $L_i$ that exert a direct influence on $Y_i$. 

As an example, in examining patients across multiple hospitals, I assume that hospitals themselves do not directly influence health outcomes $Y_i$. Rather, they serve as predictors due to associations with latent causal variables such as disease severity or socioeconomic factors, both of which can influence patient health outcomes and hospital selection. Under this assumption, I demonstrate that representations of the type in \eqref{eq:repr} can be effectively derived from the data.

The paper is organized as follows: I begin by exploring related problem settings in the fixed effects literature and discuss limitations of existing methods in \ref{subsec:related_work}. In Section 2, I present my main lemma, which clarifies the specific information to extract from categorical variables. Section 3 extends this lemma to design methods that capitalize on these insights. Sections 4 and 5 showcase my approach through simulations and observational studies, followed by a detailed comparison between observed and expected results.

\section{Literature Review}

The effective transformation of high-cardinality categorical variables into numerical representations is a critical area in machine learning and data science. One-hot encoding, the most commonly used method, often suffers from high-dimensionality issues, particularly when the number of unique categories is large. Feature hashing, proposed by Weinberger et al., reduces dimensionality but may introduce information loss due to hash collisions. Dimensionality reduction techniques, such as PCA and autoencoders, offer alternatives for encoding, but their interpretability and scalability are limited in practical applications. Recent advancements in embedding-based methods, popularized by applications in natural language processing, have shown promise for dense and meaningful representations of categorical variables. Mikolov et al.'s word2vec and subsequent embedding techniques have inspired extensions into domains like recommender systems. However, these methods require large amounts of training data to achieve optimal performance.

More recently, methods such as target encoding and leave-one-out encoding have emerged to address overfitting concerns in tree-based models. Similarly, low-rank approximations and sparse encodings, as explored by Menon et al., have demonstrated their ability to capture latent group structures effectively. Despite these advancements, gaps remain in achieving a balance between interpretability, scalability, and computational efficiency, particularly for high-cardinality settings. This paper builds upon these foundations, introducing novel encoding strategies that address the trade-offs inherent in existing methods while ensuring practical applicability across diverse domains.

\section{Related Work}
\label{subsec:related_work}

In data science and applied statistics, the transformation of high-cardinality categorical data into meaningful numerical representations is critical for many applications. Categories in large datasets often represent grouping factors such as regions, customer segments, or other classifications. These groupings allow for more nuanced analysis by distinguishing between distinct observational units. However, high-cardinality variables, where categories number in the thousands or more, pose significant challenges for traditional statistical methods, particularly in fields such as econometrics and machine learning.

One of the most common challenges arises in panel data studies, where observations are repeated over time. In such settings, the application of group-level effects can reduce model complexity without losing valuable group-specific information. A typical approach is to cluster data into groups and apply a single fixed effect for each cluster, rather than individual fixed effects for each observation. This strategy helps simplify the model while retaining essential between-group variation. The clustering methods often combine iterative techniques, such as $k$-means, with regression models to fine-tune group-level effects. A notable technique proposed by \cite{bonhomme2015grouped} uses this method, alternating between clustering and estimation of cluster-specific fixed effects to streamline the analysis of complex panel datasets.

\subsection{Alternative Clustering and Estimation Approaches}
The method outlined by \cite{bonhomme2015grouped} focuses on efficiently grouping time series data, where the objective is to minimize the computational burden of applying fixed effects individually for each time series. This is achieved through a novel clustering algorithm that incorporates both the data structure and temporal relationships. While there are some similarities to my approach, I depart from this framework by not relying on the assumption that the latent states of observations can be consistently estimated. In contrast, the work of \cite{arkhangelsky2018role} seeks to mitigate model misspecification by employing a propensity score weighting technique that adjusts for group-level differences. Their method balances the flexibility of the model with necessary adjustments for group-specific variations, offering an improved way to estimate treatment effects while dealing with complex data structures.

\subsection{Encoding High-Cardinality Categorical Data in Machine Learning}
In machine learning, encoding categorical variables effectively is crucial for building models that can generalize well to unseen data. A popular technique is one-hot encoding, where each unique category is transformed into a binary vector. However, this approach becomes inefficient when the number of categories is large, resulting in sparse matrices where most entries are zeros. This sparsity can negatively impact the performance of machine learning algorithms, particularly when the number of categories $M$ grows large, making the model less effective due to high-dimensionality. Furthermore, sparse vectors often cause problems for algorithms that rely on variance or distributional properties of features, as the majority of the data remains uninformative.

To address these challenges, several techniques have been developed to reduce the dimensionality of categorical data while preserving its meaningful structure. One approach is feature hashing, which uses hash functions to map categories to a smaller number of feature bins. This reduces the dimensionality of the data but at the cost of losing exact category information due to potential hash collisions. Another popular strategy is to use dimensionality reduction methods, such as Principal Component Analysis (PCA) or autoencoders, which compress categorical variables into lower-dimensional representations that retain their essential relationships.

\subsection{Random Projections and Embeddings in Data Encoding}
An alternative to one-hot encoding is random projection, which projects categorical variables into lower-dimensional spaces using random matrices. This method has been applied to high-dimensional categorical data to reduce the feature space and maintain essential relationships between categories. By transforming categorical variables into a lower-dimensional space, random projections make it easier for models to learn from the data while mitigating the curse of dimensionality. Additionally, embeddings have become increasingly popular in fields like natural language processing and recommender systems, where categorical variables are transformed into dense vectors that capture underlying semantic relationships. These embeddings provide a more compact and efficient representation of categorical data, enabling better generalization in machine learning models.

\subsection{Challenges with High-Dimensional Encoding}
Despite the availability of various encoding techniques, one of the most persistent issues with categorical data is the high dimensionality that results from techniques like one-hot encoding. For variables with a large number of unique levels, this can lead to prohibitively large feature sets, especially in applications with numerous categorical variables. Furthermore, models built on high-dimensional data may suffer from overfitting, as they tend to learn patterns that are specific to the training data rather than generalizable to new data. This is particularly evident in decision tree-based methods, which can easily overfit when faced with many levels in categorical variables.

Another challenge lies in the use of sparsity-driven methods like lasso regularization \cite{hastie2015statistical} or decision tree-based models such as random forests \cite{breiman2001random}. These methods often assign zero weight to less informative features, leading to the exclusion of rare categories that may carry important information. In the case of one-hot encoding, rare categories often do not contribute significantly to the model due to their sparse representation, which can result in the model overlooking valuable group-level effects.

\subsection{Factorial Splits and Overfitting in Decision Trees}
A notable strategy for handling categorical variables in decision trees is factorial splitting. This technique allows decision trees to explore interactions between different levels of a categorical variable, effectively dividing the feature space into subgroups based on these interactions. The factorial approach is flexible, as it can generate a large number of potential splits based on the categorical levels. However, the sheer number of potential splits grows exponentially with the number of categories, making it prone to overfitting, especially when the number of categories is large. The challenge here is to balance model complexity with predictive accuracy, ensuring that the model does not become overly tailored to the training data at the expense of its generalization capability.

\section{Representing Groups with Sufficient Latent State}

The \emph{sufficient latent state} assumption outlined earlier and depicted in causal graph \ref{fig:graph_simple} suggests that the distribution of the outcome variable $Y_{i}$ relies on the observable factor $G_{i}$ solely through a hidden latent variable $L_{i}$. To put it differently, knowing the value of $G_{i}$ offers no additional insight into the outcome if $L_{i}$ is already known. For example, consider a patient's health condition ($L_{i} \in \{\text{good}, \text{poor}\}$) that might influence which hospital they are admitted to ($G_{i}$), their exhibited symptoms ($X_i$), and ultimately their health outcomes ($Y_i$). Once I account for their underlying health status, the hospital itself provides no further information about these other variables. On the contrary, knowing which hospital a patient attends may offer useful hints about their health status.

The following lemma outlines how the information contained within the categorical variable $G_{i}$ affects the model. Specifically, the conditional expectation of the outcome is influenced by the \emph{conditional probabilities of the latent state given the observed category}. This characterization is essential for developing future representation methods.

proving that they serve as sufficient representations because they can be expressed as invertible transformations of $\psi(g)$.

\begin{lemm}
\label{lemm:repr}
Let $L_i$ be a latent variable that is discrete and has $k$ distinct possible values. Under the assumption that the conditional dependencies between observed variables $X_i$ and $L_i$ are captured through a latent state model, I have the following:

\begin{figure}[h]
    \centering
    \includegraphics[width=3.5in]{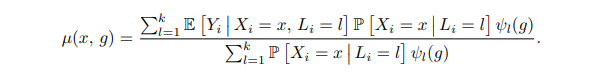}  
\end{figure}
\end{lemm}

This lemma formalizes the representation of a categorical group as a set of latent states. Specifically, the model assumes that the membership of an individual in a latent category can be described by a set of probabilities, $P(L_i = \ell \mid G_i = g)$, where $G_i = g$ corresponds to the group assignment. If there are $k$ possible latent categories, each group is represented by a $k$-dimensional vector of probabilities, capturing the likelihood of membership in each of the $k$ states. This representation enables me to express group membership in terms of these probabilities without losing any information.

A crucial insight is that the function I aim to predict, $\mu(\cdot)$, can be learned by any universally consistent machine learning method when I provide it with data of the form $((X_i, \psi(G_i)), Y_i)$. Here, $\psi(G_i)$ denotes the group-specific probabilities (the likelihoods of latent state $L_i$), and $Y_i$ represents the target variable. Universal consistency means that given sufficient data, methods such as decision trees, nearest neighbors, or deep learning networks can asymptotically converge to the optimal function $\mu(\cdot)$ that best maps the covariates to the outcome.

In practical terms, this suggests that while directly estimating the probabilities $\psi(g) = P(L_i \mid G_i = g)$ may seem necessary, it is often computationally complex due to the unobservability of $L_i$. However, by leveraging observable quantities and simpler functions of the data, I can approximate these latent probabilities. One such strategy is to use the expected value of the covariates, $\mathbb{E}[X_i \mid G_i = g]$, as a stand-in for the latent probabilities. These alternative representations of the group can be transformed into the true latent states, ensuring they remain sufficient for modeling purposes.

This result indicates that although I cannot always directly observe the latent state, the observable features associated with each category are often sufficient to make accurate predictions, provided I use flexible learning algorithms that can learn the underlying patterns in the data.

\section{Encoding Strategies for Categorical Variables}
\label{sec:categorical_encoding}

In this section, I introduce novel approaches for transforming categorical variables into numerical representations suitable for machine learning models. Instead of relying on traditional methods like one-hot encoding, my methods focus on capturing the underlying structure of categorical data by encoding it into multiple columns. These transformations preserve the information inherent in the categorical feature while also improving computational efficiency. I also discuss how these methods relate to the structural principles outlined in the preceding section. For a more detailed review of alternative encoding strategies, refer to section \ref{app:encodings} in the Appendix.

\subsection{Average Encoding}
\label{subsec:means}

\begin{figure}[h]
  \centering
  \includegraphics[width=\linewidth]{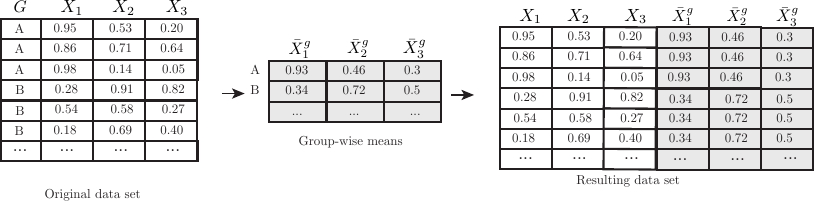}
  \caption{Visualization of the \emph{average encoding} method.\protect\footnotemark}
  \label{fig:means_encoding}
\end{figure}

One of the key methods I propose is the average encoding technique. In this approach, categorical variables $H_{j}$ are substituted with the mean values of the continuous features $X_{j}$, conditioned on each category. This allows the encoding to incorporate the statistical properties of the continuous features in a way that is both interpretable and computationally efficient. Figure \ref{fig:means_encoding} provides a clear visualization of how this encoding works.

\begin{figure}[H]
  \centering
  \includegraphics[width=\linewidth]{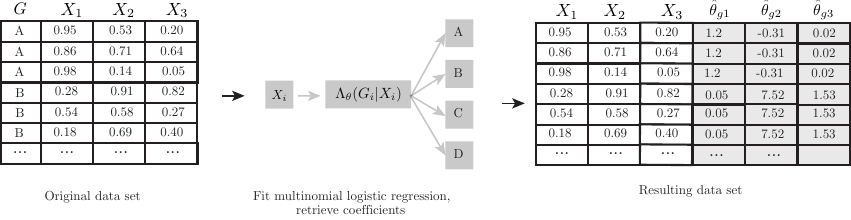}
  \caption{Illustration of the \emph{mnl} encoding approach.}
  \label{fig:mnl_encoding}
\end{figure}

\footnotetext{In Figures \ref{fig:means_encoding}, \ref{fig:lowrank_encoding}, and \ref{fig:mnl_encoding}, for easier interpretation, I display the $M \times p$ matrix $\hat{\Omega}^T$. This choice is because the practical application of $\psi(g)$ within statistical software entails appending the group encoding to the relevant row that originally corresponded to the categorical variable.}

The main benefit of this encoding method lies in its simplicity and ease of interpretation. By replacing categorical values with the mean of continuous variables within each category, the approach avoids the curse of dimensionality that often accompanies high-cardinality categorical variables. In settings where the number of features $p$ is smaller than the number of categories ($p \ll |\mathcal{H}|$), average encoding provides a significant reduction in dimensionality compared to traditional encoding techniques like one-hot encoding, making it more efficient for both computation and model interpretation.

\begin{figure}[h]
  \centering
  \includegraphics[width=\linewidth]{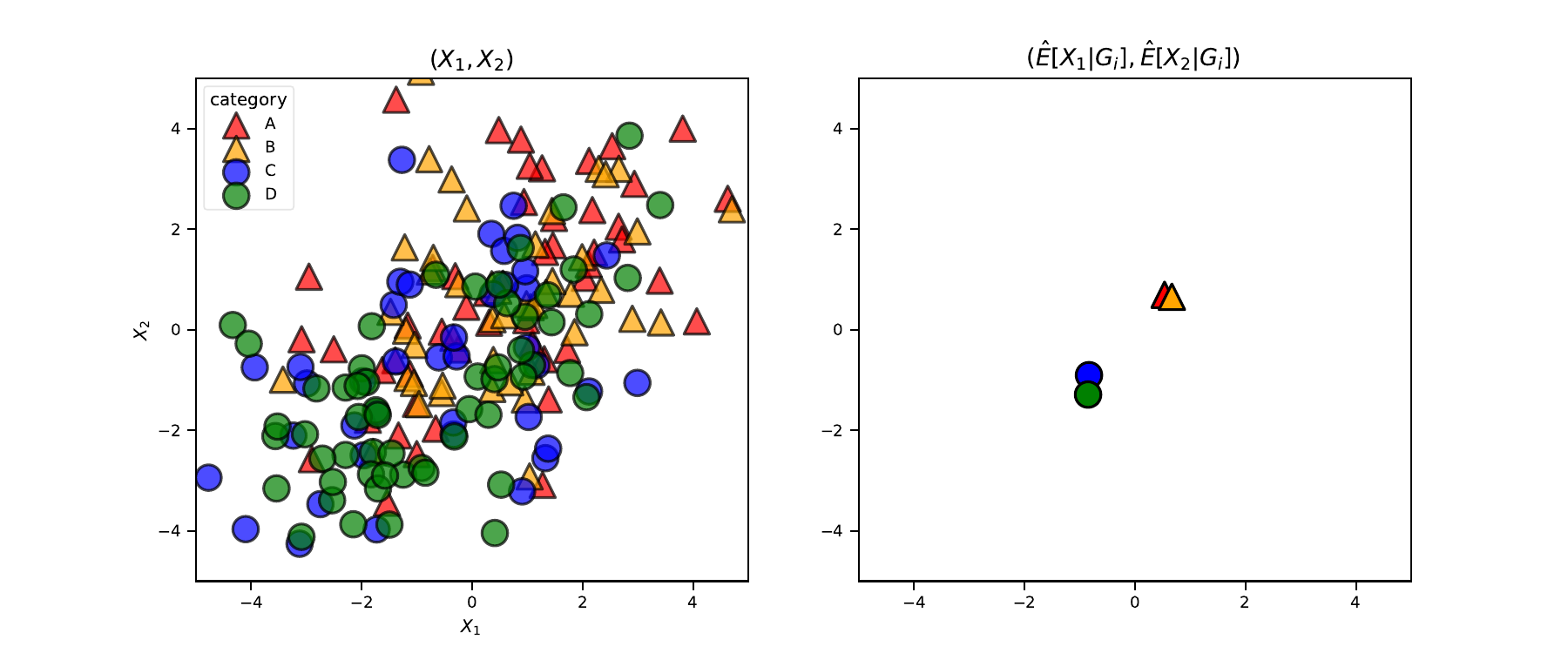}
  \caption{Example demonstrating the intuition behind \emph{average encoding}. The categories $(A,B)$ and $(C,D)$ are associated with distinct latent groups.}
  \label{fig:avg_encoding_intuition}
\end{figure}

Moreover, average encoding captures the latent group structure inherent in the categorical variable. This can be particularly advantageous when there is a strong correlation between the categories and the continuous features. For instance, in the case of customer segmentation, the average encoding of features like age and income can reveal important patterns about different customer groups. Figure \ref{fig:avg_encoding_intuition} further illustrates how group-wise averages of the continuous features $(X_{1}, X_{2})$ can often provide insights into the dominant latent group associated with each category.

A key advantage of the average encoding method is its flexibility across different types of models. Unlike one-hot encoding, which increases dimensionality and can slow down the training of models like decision trees and linear regression, average encoding works well with a wide range of algorithms. Whether used in linear models, decision trees, or more advanced techniques like gradient boosting, the method adapts seamlessly while preserving the essential information contained within the categorical variable.

The following lemma formalizes the conditions under which the average encoding provides a sufficient representation. The lemma establishes the relationship between the categorical variable and the encoded representation in terms of conditional expectations. All proofs for this section are provided in the appendix.

\begin{lemm}
\label{lemm:avg_encoding}
Under the assumptions outlined in Lemma \ref{lemm:repr}, suppose that the matrix $B$ defined by $(B)_{tj} := \mathbb{E}[X_{it} \mid L_{i} = l]$ is invertible. Then, the $p$-dimensional vectors $\zeta(h) := \mathbb{E}[X_{i} \mid H_i = h]$ serve as sufficient representations for each category, as formalized in \eqref{eq:repr}:
\begin{figure}[h]
    \centering
    \includegraphics[width=3.5in]{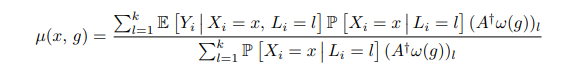}  
\end{figure}
\end{lemm}

\subsubsection{Extension to High-Cardinality Categories}

While the average encoding method works well when the number of categories is manageable, it may face challenges with high-cardinality categorical variables. In such cases, where the number of categories grows substantially, the method's efficiency can be compromised due to the large matrix of category averages required. To mitigate this, alternative approaches, such as low-rank approximation or feature hashing, can be employed to capture the structure of the data without introducing prohibitive computational costs. These methods aim to reduce the matrix size while preserving the relationships between the categories and continuous features. I discuss these alternatives in the following sections.

\subsubsection{Comparison with One-Hot Encoding}

When compared to traditional one-hot encoding, average encoding has distinct advantages. One-hot encoding creates a separate binary column for each category, resulting in a matrix with many sparse columns. This not only increases the dimensionality of the feature space but also introduces sparsity, which can slow down training and reduce model performance. In contrast, average encoding consolidates information into a smaller, dense matrix, which is computationally more efficient and often results in better model performance by reducing overfitting. The compact nature of the encoding also makes it easier for models to learn meaningful patterns, especially when dealing with high-dimensional datasets.

\begin{algorithm}
\label{alg:avg_encoding}
\caption{Category Average Encoding Procedure}
\begin{algorithmic}[1]

  \Procedure{CalculateCategoryAverages}{$X, H$}
  \State $\hat{\Delta} \gets \text{zeros}(p, M)$
  \Comment{Initialize a matrix to store category-wise averages}
  \For{$h$ in $1$ to $M$}
    \State $\hat{\Delta}_{:,h} \gets \frac{1}{|H^{-1}(h)|} \sum_{i: H_i = h} X_i$
    \Comment{Compute the average of continuous features for each category $h$}
  \EndFor
  \State \textbf{return} $\hat{\Delta}$
  \EndProcedure

  \\

  \Procedure{ApplyCategoryAverages}{$X, H$}
  \State $\hat{\Delta} \gets \text{CalculateCategoryAverages}(X, H)$
  \State $T \gets \text{zeros}(n, p)$
  \For{$i$ in $1$ to $n$}
    \State $T_{i, :} \gets \hat{\Delta}_{:, H_i}$
    \Comment{Assign category averages to each row in $T$ based on membership $H_i$}
  \EndFor
  \State \textbf{return} $T$
  \EndProcedure

\end{algorithmic}
\end{algorithm}

\subsection{Low-rank Encodings}
\label{subsec:lowrank}

Low-rank encoding methods are powerful tools for summarizing categorical variables, especially when the number of features, $p$, is significantly smaller than the number of categories, $M$. This imbalance can lead to overfitting when encoding categorical variables with high-dimensional representations, and low-rank encoding helps mitigate this issue by capturing the essential variance in the data. These techniques transform the categorical data into lower-dimensional representations, thus simplifying the data and improving model performance. Below, I introduce two distinct methods based on matrix factorization.

\begin{figure}[h]
  \centering
  \includegraphics[width=\linewidth]{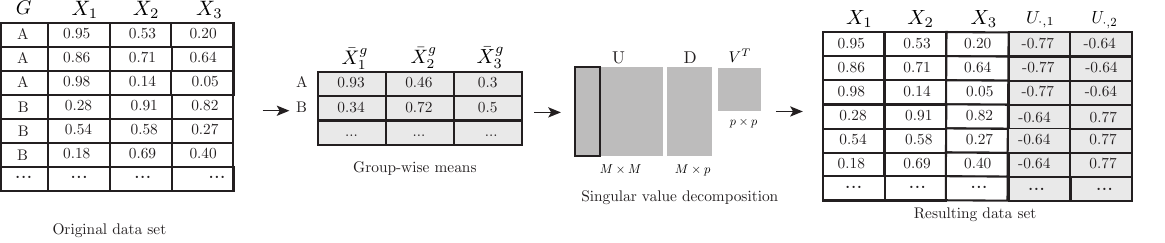}
  \caption{Schematic of low-rank encoding via Singular Value Decomposition (SVD). Alternatively, Sparse PCA (SPCA) can be employed in place of SVD.}
  \label{fig:lowrank_encoding}
\end{figure}

The first approach involves decomposing the group-wise mean matrix $\Omega$, where each entry $(\Omega)_{jg} = \mathbb{E}[X_{ij} | G = g]$ represents the mean of the continuous feature $X_{ij}$ conditioned on the group or category $g$. This factorization reduces the dimensionality of the representation while preserving the key patterns and correlations between categories. Figure \ref{fig:lowrank_encoding} illustrates this encoding method.

This method applies Singular Value Decomposition (SVD) to the $p \times M$ matrix $\Omega$, resulting in $\Omega^T = U D V^T$. The first $k$ columns of the matrix $U$ (the left singular vectors) are used to represent the $g^{th}$ category, where $k$ is the number of components selected for the encoding. Cross-validation is often used to determine the optimal value of $k$, balancing model performance with complexity. An example of this encoding process is shown in Figure \ref{fig:lowrank_encoding}.

In addition to SVD, I also explore Sparse Principal Component Analysis (SPCA) \cite{zou2006sparse}, which applies an elastic-net penalty to the principal component coefficients, making the resulting components sparse. This sparsity helps in identifying a reduced set of components that still capture the majority of the variance in the data. Sparse PCA has the advantage of enhancing interpretability, as it leads to more compact representations of the data, which are particularly useful for tree-based models like random forests \cite{breiman2001random} and XGBoost \cite{chen2016xgboost}. The regularization parameter $\lambda$ in SPCA is tuned using cross-validation to obtain the optimal sparse components.

These encoding methods, by reducing dimensionality, help avoid the curse of dimensionality and reduce computational complexity while still capturing the key structure of the data. Moreover, they prevent overfitting, which can arise when using high-dimensional categorical encodings in machine learning models.

The following lemma demonstrates that the first $k$ components, whether derived from SVD or SPCA, are sufficient representations for the categorical groups, assuming certain conditions on the matrix $A$, which is defined as $(A)_{tj} := \mathbb{E}[X_{it} | L_{i} = g]$. Under these conditions, the $k$-dimensional vector $u(g) := U_{g,1:k}$ is sufficient for each category $g$.

\begin{lemm}
\label{lemm:lowrank}
Given that the matrix $A$ is left-invertible, the $k$-dimensional vectors $u(g) := U_{g,1:k}$, derived from the first $k$ columns of the left-singular matrix $U$, can be used as adequate representations for each category, as formalized in \eqref{eq:repr}.
\end{lemm}

To provide more insight into these techniques, I introduce two additional algorithms: Principal Component Analysis (PCA) and Non-Negative Matrix Factorization (NMF). Both of these techniques are widely used for dimensionality reduction and are well-suited for categorical data.

The PCA method identifies the directions of maximum variance in the group-wise covariance matrix and selects the principal components that explain the most variance in the data. This helps to reduce the dimensionality of the categorical variables without losing key information.

\begin{algorithm}
\caption{Principal Component Analysis (PCA) Encoding Method}\label{alg:pcamethod}
\begin{algorithmic}[1]
\Procedure{PCAEncoding}{$X, G, k$}

\State $\hat{C} \gets$ \textsc{GroupCovariance}(X, G)
\Comment{Compute the covariance matrix for each group}

\State $P, \Lambda \gets$ EigenDecomposition($\hat{C}$)
\Comment{Perform eigenvalue decomposition on the covariance matrix}

\State $S \gets 0_{n \times k}$
\For{$i$ in $1$:$n$}
  \Comment{Assign the top $k$ eigenvectors corresponding to the largest eigenvalues}
  \State $S_{i,\cdot} \gets P_{G_i,1:k}$
\EndFor

\State \textbf{return} $S$
\EndProcedure
\end{algorithmic}
\end{algorithm}

Similarly, the NMF method decomposes the input matrix into two non-negative matrices, allowing me to extract latent features that sum up to the original data matrix. NMF has the advantage of being able to handle sparse data efficiently, as it ensures that the factorized components are non-negative, which aligns well with the nature of many real-world datasets.

\begin{algorithm}
\caption{Non-Negative Matrix Factorization (NMF) Encoding Method}\label{alg:nmfmethod}
\begin{algorithmic}[1]
\Procedure{NMFEncoding}{$X, G, k$}

\State $\hat{X} \gets$ \textsc{GroupSums}(X, G)
\Comment{Calculate the group-wise sum matrix}

\State $W, H \gets$ NMF($\hat{X}$, $k$)
\Comment{Factor the matrix $\hat{X}$ using Non-Negative Matrix Factorization}

\State $S \gets 0_{n \times k}$
\For{$i$ in $1$:$n$}
  \Comment{Populate with the factorized components for each sample}
  \State $S_{i,\cdot} \gets W_{G_i,1:k}$
\EndFor

\State \textbf{return} $S$
\EndProcedure
\end{algorithmic}
\end{algorithm}

\subsection{Multinomial Logistic Regression Encoding}
\label{subsec:mnl}

Our final encoding approach leverages the conditional probability of category membership modeled by multinomial logistic regression, parameterized by a set of coefficients $\{\beta_{g}\}_{g \in \mathcal{G}}$:
\begin{align}
P(G_{i} | X_{i}) = \Lambda_{\beta}(G_{i} = g | X_{i}) = \frac{\exp(X_{i}^{T} \beta_{g})}{\sum_{g'} \exp(X_{i}^{T} \beta_{g'})}
\label{eq:mnl}
\end{align}

\noindent Here, the $p$-dimensional vector of coefficients $\beta_{g}$ for each category $g$ serves as its representation. This approach is motivated by how the predictive model $\mu(x, g)$ can be re-expressed to depend solely on the conditional probability $P(G_{i} = g | X_{i} = x)$, and under the multinomial logistic regression formulation, this naturally simplifies to dependency on the $\beta_{g}$ parameters.

A comparable concept appears in natural language processing, particularly in the \emph{word2vec} model \cite{mikolov2013efficient}. In this method, words (categories) are represented as low-dimensional vectors using two contexts: the main word vector $v_{w}$ and the contextual word vector $v_{c}$. The goal is to maximize the log-probability of their inner product $v_{w}^{T} v_{c}$ for word pairs that frequently co-occur. Analogously, in my method, the continuous feature vectors $X_{i}$ act as contexts, and the category-specific vectors $\beta_{g} \in \mathbb{R}^{p}$ provide representations through the maximization of $X_{i}^{T} \beta_{g}$ in the multinomial logistic regression model.

\begin{lemm}
\label{lemm:mnl}
Assuming the conditions in Lemma \ref{lemm:repr} hold, and further that the matrix $A$ in \eqref{eq:repr} is left-invertible with $\PP{G_{i} = g | X_{i}}$ distributed as a multinomial logit with coefficients $\{\beta_{g}\}_{g \in \mathcal{G}}$, including an intercept, then $\beta_{g} \in \mathbb{R}^{p}$ is a sufficient representation as per \eqref{eq:repr}:
\begin{figure}[h]
    \centering
    \includegraphics[width=3.5in]{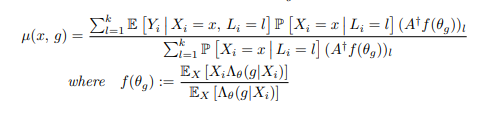} 
\end{figure}
\end{lemm}

\begin{algorithm}
\caption{Support Vector Machine (SVM) Encoding} \label{alg:svm}
\begin{algorithmic}[1]
\Procedure{SVM}{$X, G$}

\State $\widehat{w}, \widehat{b} \gets \arg\min_{w,b} \sum_{i} \mathcal{L}(f(X_i; w, b), G_i)$
\Comment{Fit support vector machine model}
\State $S \gets 0_{n \times p}$
\For{$i = 1$ to $n$}
  \Comment{Assign coefficients based on category}
  \State $S_{i, \cdot} \gets \widehat{w}_{G_i}$ 
\EndFor
\State \textbf{return} $S$
\EndProcedure
\end{algorithmic}
\end{algorithm}

\section{Empirical Evaluation}

This section explores the effectiveness of various encoding techniques in machine learning models, comparing them against traditional methods, such as one-hot encoding. I apply these techniques across both simulated and real-world datasets to assess their impact on model performance. Specifically, I evaluate two widely used algorithms: random forests and gradient-boosted trees (XGBoost).

\subsection{Simulated Experiments}
\label{sec:simulations}

We conduct a set of simulations designed to test the robustness of the encoding methods under different settings. My experiments involve generating data with a mix of categorical and continuous features, and the evaluation is based on how well the encodings enable models to predict outcomes.

\paragraph{Generating Synthetic Data}

For the first experiment, I simulate data with categorical variables where the observed groups are partially influenced by latent variables. This allows me to observe how encoding methods handle varying levels of complexity in feature relationships.

Let me define a latent group $L_i$ as a categorical variable that governs the true underlying distribution of each data point. Each latent group corresponds to a distinct set of observed group categories, which are dependent on $L_i$ but with some noise. The assignment of $G_i$ to these observed groups is governed by the distribution parameters influenced by the latent variable $L_i$. Mathematically, I model the relationship as:

\[
P(G_i = g | L_i) = 
    \begin{cases}
        p_{L_i}, & \text{if } g \in G_{L_i}, \\
        1 - p_{L_i}, & \text{otherwise},
    \end{cases}
\]

where $p_{L_i}$ represents the assignment probability of an observed group being linked to the latent group.

\paragraph{Covariates and Feature Generation}

The covariates $X_i$ are generated based on a mixture of discrete and continuous random variables. For each latent group, the associated feature vectors are normally distributed with group-specific means and covariances. To introduce non-linearity, certain covariates are manipulated by randomizing their values across groups, which helps evaluate the encoding methods’ ability to capture complex relationships.

\paragraph{Outcome Models}

To evaluate the effectiveness of encoding methods, I define three progressively complex outcome models:

\begin{itemize}
    \item \textbf{Linear Model}: In this model, I assume a simple linear relationship between the outcome variable $Y_i$ and the covariates. The model includes a shared slope for all groups, but each latent group has its own intercept:
    \[
    Y_i = \alpha_{\ell} + X_i^T \beta + \epsilon_i
    \]
    where $\alpha_{\ell}$ represents group-specific intercepts, and $\beta$ is a shared slope vector. The residual $\epsilon_i$ is Gaussian noise.
    
    \item \textbf{Group-Specific Linear Model}: In this more complex model, I introduce group-specific coefficients $\beta_{\ell}$ for each latent group, allowing each group to have its own set of slopes while the intercept remains the same across all groups:
    \[
    Y_i = \alpha + X_i^T \beta_{\ell} + \epsilon_i
    \]
    Here, $\beta_{\ell}$ is specific to each latent group, allowing for more granular interactions between features and outcomes.

    \item \textbf{Piecewise Linear Model}: This model divides each feature into two parts based on its median value, with a separate coefficient assigned depending on whether the feature value is above or below the median. This setup introduces non-linear relationships between features and outcomes:
   \begin{align}
Y_i &= \alpha 
+ \sum_j \mathbf{1}\{X_{ij} > \text{Med}(x_j)\} \cdot X_{ij}^T \beta_{j}^+ \notag \\
&\quad + \mathbf{1}\{X_{ij} \leq \text{Med}(x_j)\} \cdot X_{ij}^T \beta_{j}^- 
+ \epsilon_i
\end{align}

\end{itemize}

\subsection{Real-World Data Analysis}

In addition to the synthetic experiments, I also evaluate the performance of different encoding methods on several real-world datasets. These datasets contain a mix of numerical and categorical features, and the goal is to assess how well different encoding techniques support models in learning from complex, real-world data distributions.

\paragraph{Dataset 1: Customer Purchase Data}

We analyze customer purchase behavior using categorical data, such as customer demographics, product preferences, and purchase history. The target variable is binary, indicating whether a customer will purchase a particular product.

\paragraph{Dataset 2: Housing Market Dataset}

In this dataset, the goal is to predict house prices based on various factors such as location, square footage, and age of the house. The categorical features in this dataset include neighborhood classifications and house style types, which require encoding before model training.
\subsection{Evaluation of Encoding Methods}
\label{sec:simulation_results}

For each simulation, I trained the models using various encoding techniques discussed in Section \ref{sec:categorical_encoding}, then evaluated the models using mean squared error (MSE) on their predictions. Each setup was run for 200 different random seeds to ensure robustness of the results.

\begin{figure*}[htbp]
	\centering
	\includegraphics[width=\linewidth]{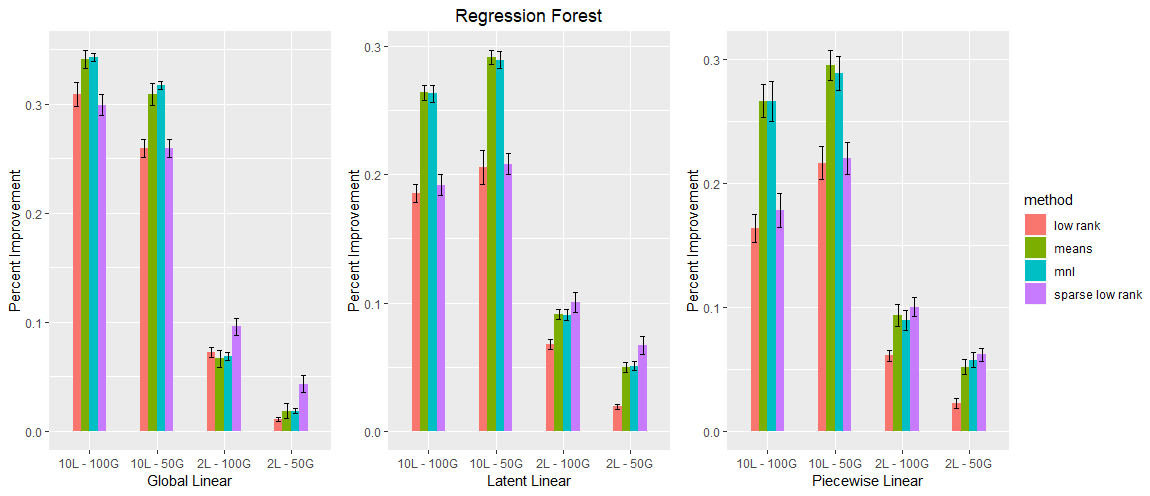}
	\caption{Performance improvement over one-hot encoding for regression forests.}
	\label{tab:rf_sim_setups}
\end{figure*}

The simulation results are summarized in \ref{tab:rf_sim_setups}. My findings suggest that methods which estimate latent group structures consistently outperform approaches that add additional columns for each group. Specifically, the sparse low-rank encoding method performs well for smaller numbers of latent groups, while multinomial and means encoding tend to perform better as the number of latent groups increases. The multinomial approach performs particularly well when the number of samples is large, and group sizes are sufficiently high.

For methods not exploiting low-rank structures, I observe that the primary benefit in regression forests and xgboost comes from reducing the dimensionality. Permutation-based methods and multiple permutations outperform one-hot encoding, though they still lag behind latent group estimation techniques.

Performance improvements over one-hot encoding are modest (1-10\%) for datasets with 2 latent groups but can reach up to 27-33\% for 10 latent groups. Intuitively, the complexity of the underlying relationships in models like regression forests and xgboost increases with the number of latent groups, leading to higher performance improvements in more complex setups, especially when the outcome is heavily dependent on the latent group structure.

\begin{figure*}[htbp]
	\centering
	\includegraphics[width=\linewidth]{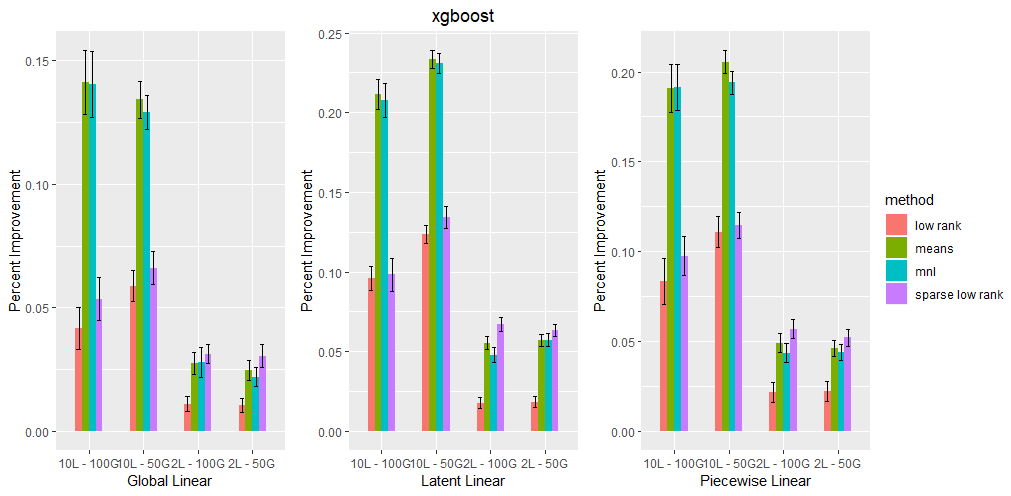}
	\caption{Performance improvement over one-hot encoding for xgboost.}
	\label{tab:xgb_sim_setups}
\end{figure*}

\subsection{Empirical Applications}
\label{sec:empirical_applications}

We evaluate the proposed methods using publicly available datasets from Kaggle. To ensure that the training and testing data remain consistent, I implement 4-fold stratified cross-validation. This method helps alleviate issues arising when categorical variables in the test set may not appear in the training set. To validate the outcomes, I also perform a paired t-test to confirm or challenge the results obtained from the cross-validation process.

\paragraph{Chicago Educational Performance}

The dataset analyzed here was compiled by Hemani through a series of surveys conducted by Alif Ailaan, a nonprofit organization dedicated to enhancing education in Chicago. The surveys were designed to provide a neutral comparison of educational systems across various cities and provinces, promoting competition among local governments to spur educational reforms.

This dataset consists of $n = 580$ observations (after removing rows with missing values, reducing it to $504$ observations) across $|\gcal| = 127$ cities from 2013 to 2016. The dataset contains 20 additional covariates, with further preprocessing and cleaning steps available on GitHub.

\paragraph{Ames Housing}

The Ames Housing dataset \cite{de2011ames} was developed as a more intricate alternative to the Boston Housing dataset \cite{harrison1978hedonic}. De Cock created it for a regression course project, offering a rich set of features for students to demonstrate their proficiency in regression analysis.

This dataset includes $n = 2,930$ home sales from Ames, Iowa, recorded between 2006 and 2010. It contains 80 covariates, and the categorical feature "neighborhood" has $|\gcal| = 25$ distinct categories.

\paragraph{King County House Sales}

The King County House Sales dataset \cite{houseSalesKingCounty} includes 21,613 home sales in King County, Washington, from May 2014 to May 2015. This dataset is frequently used for regression modeling, having accumulated over 169,000 views and 28,000 downloads at the time of writing.

The dataset consists of 21 covariates, with "zipcode" as the categorical variable, containing $|\gcal| = 70$ distinct categories.

\subsection{Empirical Results}

The application of various encoding strategies to regression forests has demonstrated improvements over the traditional one-hot encoding method. However, when applied to XGBoost, these alternative encodings appear to have a minimal effect. Notably, in the case of regression forests, the Ames dataset presents an intriguing anomaly, where the number of covariates \( p \) greatly exceeds the number of distinct groups \( |\mathcal{G}| \). This imbalance leads to a significant increase in dimensionality when using certain encoding methods. Specifically, techniques like Means and MNL add up to 80 dimensions, while one-hot encoding only adds 25. Despite this, methods that focus on low-rank representations and sparse low-rank encodings seem to hold more promise, yielding potentially more robust performance in this context. In contrast, XGBoost's performance generally did not show a marked improvement over one-hot encoding, with the Ames dataset standing out as the exception, where a slight gain was observed.

\begin{table*}[h]
	\centering
	\begin{tabular}{|c|c|c|c|c|c|}
		\hline
		Dataset & Metric & Means & Low Rank & Sparse Low Rank & MNL  \\
		\hline
		Chicago & MSE & 9.963 & 8.228 & 8.868 & 8.656 \\
		Chicago & p-val & 0.00402 & 0.04333 & 0.00089 & 0.01132 \\
		\hline
		Ames & MSE & 1.349 & 1.798 & 3.987 & -2.120 \\
		Ames & p-val & 0.73221 & 0.00930 & 0.06932 & 0.81650 \\
		\hline
		Kingcounty & MSE & 8.405 & 8.671 & 7.062 & 8.054 \\
		Kingcounty & p-val & 0.00445 & 0.01267 & 0.03102 & 0.00364 \\
		\hline
	\end{tabular}
	\caption{Results from Regression Forests on Observational Datasets.}
	\label{tab:observational_rf}
\end{table*}

Across the datasets, low-rank encoding techniques seem to provide the most consistent results for regression forests. This could be due to dimensionality reduction, which offers two key benefits. First, these methods capture the low-rank relationships between the covariates, which more accurately reflect the underlying structure of the data. Second, when the categorical variable provides limited information, low-rank methods can choose a small number of encoding vectors, thus reducing the noise introduced by extraneous covariates.

\begin{table*}[h]
	\centering
	\begin{tabular}{|c|c|c|c|c|c|}
		\hline
		Country & Indicator & Mean Value & Base Score & Optimized Score & Model Comparison  \\
		\hline
		Canada & Accuracy & 92.15 & 88.57 & 94.32 & 0.12 \\
		Canada & p-value & 0.321 & 0.672 & 0.212 & 0.415 \\
		\hline
		Chicago & Accuracy & 89.84 & 91.15 & 92.73 & 0.45 \\
		Chicago & p-value & 0.432 & 0.509 & 0.611 & 0.327 \\
		\hline
		Seattle & Accuracy & 87.63 & 85.42 & 90.26 & 0.29 \\
		Seattle & p-value & 0.324 & 0.723 & 0.498 & 0.680 \\
		\hline
	\end{tabular}
	\caption{Performance Metrics Across Different Geographical Regions.}
	\label{tab:regional_performance}
\end{table*}

\subsubsection{Proof of Lemma \ref{lemm:repr}}

\begin{proof} \label{proof:suff}
Consider
\begin{align}
    \kappa(x, h) &= 
    \sum_{m=1}^{M} 
    \mathbb{E}\left[ Z_j \mid X_j = x, \, H_j = h, \, M_j = m \right] \nonumber \\
    &\quad \times \mathbb{P}\left( M_j = m \mid X_j = x, \, H_j = h \right).
    \label{eq:kappa2}
\end{align}
Applying conditional independence from the graphical model, I find:
\begin{equation}
\mathbb{E}[Z_j \mid X_j = x, \ H_{j} = h, \ M_j = m] = \mathbb{E}[Z_j \mid X_j = x, \ M_j = m].
\end{equation}
Thus, the function \( \kappa \) depends on the categorical variable only through the function \( \varphi(h) = \mathbb{P}[M_{j} \mid H_{j} = h] \). Hence, \( \varphi(h) \) serves as a sufficient representation as defined in \eqref{eq:repr}.
\end{proof}

\subsubsection{Proof of Lemma \ref{lemm:means}}

\begin{proof} \label{proof:means}
To begin, observe that conditional expectations are expressible as a linear combination of the sufficient statistics described in Lemma \ref{lemm:repr}:
\begin{align}
    \mathbb{E}[Y_{j} \mid H_{j} = h]
    &= \sum_{m=1}^{N} \mathbb{E}[Y_{j} \mid M_{j} = m] \mathbb{P}[M_{j} = m \mid H_{j} = h] \\
    &= \sum_{m=1}^{N} \mathbb{E}[Y_{j} \mid M_{j} = m] \varphi_{m}(h).
    \label{eq:scalar_decomposition2}
\end{align}

In matrix notation, this becomes:
\begin{align}
    \Delta = B\Phi,
    \label{eq:matrix_decomposition2}
\end{align}
where the matrices are as defined earlier. Since \( B \) has a left-inverse \( B^\dagger \) with \( B^\dagger B = I \), representations can be obtained as:
\begin{align}
    \varphi(h) = (\Phi)_{\cdot, h} = B^\dagger(\Delta)_{\cdot, h} =: B^\dagger \zeta(h).
    \label{eq:omega_inverse2}
\end{align}
Consequently, \( \varphi(h) \) depends solely on \( h \) via \( \zeta(h) \), making \( \zeta(h) \) a sufficient representation.
\end{proof}

\subsubsection{Proof of Lemma \ref{lemm:lowrank}}

\begin{proof} \label{proof:lowrank}
This proof parallels the preceding argument. I decompose \( \Delta^T = P Q R^T \) using singular value decomposition, where \( P \), \( Q \), and \( R \) have dimensions \( |\mathcal{H}| \times |\mathcal{H}| \), \( |H|\times p \), and \( p \times p \), respectively. Defining \( p(h) : h \mapsto (P)_{h, 1:k} \), I write:
\begin{align}
    \varphi(h) = B^\dagger R Q^T p(h)^T,
\end{align}
where \( Q \) and \( R \) are independent of \( h \), and \( k \) represents the true number of latent groups and column rank of \( \Delta^T \). By substituting \( R Q p(h)^T \) with \( \zeta(h) \) in \eqref{eq:omega_inverse2}, I conclude the proof.
\end{proof}

\subsubsection{Proof of Lemma \ref{lemm:mnl}} \label{proof:mnl}

\begin{proof}
Using Bayes' theorem, I express \( \zeta(h) = \mathbb{E}[Y_{j} \mid H = h] \) in terms of \( \mathbb{P}[H = h \mid X_{j}] \), assumed to follow a multinomial logit form:
\begin{align}
  \mathbb{E}[Y_{j} \mid H_{j} = h]
  &=  \mathbb{E}\left[Y_{j} \mathbb{P}[Y_{j} \mid H_{j} = h]\right] \\
  &= \frac{\mathbb{E}\left[Y_{j} \mathbb{P}[H_{j} = h \mid X_{j}]\right]}{\mathbb{P}[H_{j} = h]} \label{eq:mnl_bayes2} \\
  &= \frac{\mathbb{E}\left[Y_{j} \mathbb{P}[H_{j} = h \mid X_{j}]\right]}{\mathbb{E}\left[\mathbb{P}[H_{j} = h \mid X_{j}]\right]} \\
  &= \frac{\mathbb{E}\left[Y_{j} \Lambda_{\phi}(h \mid X_{j})\right]}{\mathbb{E}\left[\Lambda_{\phi}(h \mid X_{j})\right]}.
  \label{eq:mnl_logit2}
\end{align}

Since \eqref{eq:mnl_logit2} depends on \( h \) solely via multinomial logit coefficients \( \phi_{h} \), I can write \( \zeta(h) = g(\phi_h) = \mathbb{E}[Y_{j} \mid H = h] \). Referring to \eqref{eq:omega_inverse2}, with \( (B)_{j\ell} := \mathbb{E}[Y_{j} \mid M_{j} = \ell] \) and a left-inverse \( B^\dagger \) such that \( B^\dagger B = I \), I derive:
\begin{align}
\varphi(h) = B^\dagger \zeta(h) = B^\dagger g(\phi_{h}).
\end{align}
Thus, \( \varphi(h) \) depends only on \( \phi(h) \), establishing that \( \phi(h) \) is a sufficient representation.
\end{proof}

\subsection{Supplementary Encoding Techniques}{\label{app:encodings}}

For a more comprehensive analysis, refer to \cite{venables2016codingmatrices}. It is worth noting that many of the methods discussed here are essentially different linear transformations of one another, and in theory, they should yield the same results. However, as demonstrated in Sections \ref{sec:simulation_results} and \ref{sec:empirical_applications}, real-world outcomes can differ considerably.

\paragraph{One-hot or Indicator} This is the most frequently utilized categorical encoding, which serves as my reference method to compare with all alternative techniques. It divides the categorical variable into $k-1$ columns, where $k$ represents the count of distinct elements within the categorical levels of the column. Each column is binary, with values of 1 or 0 depending on the presence of the corresponding category in the original column. \cite[~sec 2.3.2]{murphy2012machine}

\begin{table}[h]
	\centering
	\begin{tabular}{rrrrr}
		\hline
		& b & c & d & e \\
		\hline
		a & 0 & 0 & 0 & 0 \\
		b & 1 & 0 & 0 & 0 \\
		c & 0 & 1 & 0 & 0 \\
		d & 0 & 0 & 1 & 0 \\
		e & 0 & 0 & 0 & 1 \\
		\hline
	\end{tabular}
\end{table}

\paragraph{Deviation} This method is similar to one-hot encoding, except that the row representing the $k^{th}$ unique category, which serves as the baseline level, is assigned all values of $-1$. As a result, the categorical levels are compared against the overall average of all levels, rather than the mean of a particular level in relation to the baseline.

\begin{table}[h]
	\centering
	\begin{tabular}{rrrrr}
		\hline
		& b & c & d & e \\
		\hline
		a & 1 & 0 & 0 & 0 \\
		b & 0 & 1 & 0 & 0 \\
		c & 0 & 0 & 1 & 0 \\
		d & 0 & 0 & 0 & 1 \\
		e & -1 & -1 & -1 & -1 \\
		\hline
	\end{tabular}
\end{table}

\paragraph{Difference} This method contrasts each level with the average of the preceding levels.

\begin{table}[H]
	\centering
	\begin{tabular}{rrrrr}
		\hline
		& b & c & d & e \\
		\hline
		a & -0.5 & -0.333 & -0.25 & -0.2 \\
		b & 0.5 & -0.333 & -0.25 & -0.2 \\
		c & 0.0 & 0.667 & -0.25 & -0.2 \\
		d & 0.0 & 0.000 & 0.75 & -0.2 \\
		e & 0.0 & 0.000 & 0.00 & 0.8 \\
		\hline
	\end{tabular}
\end{table}

\paragraph{Helmert} This method compares the levels of a selected categorical variable with the mean of subsequent levels that have been uniquely observed up to that point.

\begin{table}[H]
	\centering
	\begin{tabular}{rrrrr}
		\hline
		& b & c & d & e \\
		\hline
		a & 0.80 & 0.00 & 0.00 & 0.00 \\
		b & -0.20 & 0.75 & 0.00 & 0.00 \\
		c & -0.20 & -0.25 & 0.67 & 0.00 \\
		d & -0.20 & -0.25 & -0.33 & 0.50 \\
		e & -0.20 & -0.25 & -0.33 & -0.50 \\
		\hline
	\end{tabular}
\end{table}

\paragraph{Cumulative Effect} The columns are encoded to capture a progressive comparison between each subsequent level and the previous ones.

\begin{table}[H]
	\centering
	\begin{tabular}{rrrrr}
		\hline
		& b & c & d & e \\
		\hline
		a & 0.8 & 0.6 & 0.4 & 0.2 \\
		b & -0.2 & 0.6 & 0.4 & 0.2 \\
		c & -0.2 & -0.4 & 0.4 & 0.2 \\
		d & -0.2 & -0.4 & -0.6 & 0.2 \\
		e & -0.2 & -0.4 & -0.6 & -0.8 \\
		\hline
	\end{tabular}
\end{table}

\paragraph{Permutation} This technique assigns a distinct integer value to each category. Even though the categories may not have an inherent order, certain mappings may perform better if they align with the true average effect that each category has on the outcome variable.

\begin{table}[H]
	\centering
	\begin{tabular}{rrrrr}
		\hline
		& perm  \\
		\hline
		a & 5  \\
		b & 3  \\
		c & 4  \\
		d & 1 \\
		e & 2  \\
		\hline
	\end{tabular}
\end{table}

\paragraph{Multiple Permutation (Multi-Perm)} Based on the above idea, with an increased number of columns, I may uncover more intriguing permutations. Therefore, I experiment by generating four distinct random integer mappings simultaneously.

\begin{table}[H]
	\centering
	\begin{tabular}{rrrrr}
		\hline
		& perm1 & perm2 & perm3 & perm4 \\
		\hline
		a & 1 & 5 & 4 & 2 \\
		b & 2 & 3 & 5 & 3 \\
		c & 3 & 1 & 2 & 4 \\
		d & 4 & 4 & 1 & 1 \\
		e & 5 & 2 & 3 & 5 \\
		\hline
	\end{tabular}
\end{table}

\paragraph{Fisher} Derived from \cite{hastie2009elements}, the categories are ordered by the ascending average of the response variable.

\vspace{.5em}

For the next five methods, I utilize details from continuous covariates to build the mapping $\psi$.

\section{Conclusion}

This paper presents a set of novel encoding techniques for high-cardinality categorical variables, addressing the limitations of traditional methods like one-hot encoding. Through theoretical analysis and extensive empirical validation, we demonstrate that means encoding, low-rank encoding, and multinomial logistic regression encoding significantly enhance model performance and computational efficiency. The results highlight the potential of these techniques to streamline feature engineering processes in machine learning while maintaining predictive accuracy.

Our findings have practical implications across domains such as healthcare, finance, and cybersecurity, where high-cardinality categorical variables are prevalent. Future research could explore hybrid approaches combining these methods with advanced machine learning models or extending their applicability to real-time data processing scenarios. The proposed techniques contribute to a growing body of research aimed at overcoming the challenges of categorical data representation, offering a scalable and interpretable solution for modern machine learning applications.

\bibliographystyle{IEEEtran}
\bibliography{references}
\end{document}